\pdfoutput=1
\documentclass{article}

\usepackage[final]{neurips_2020_tda}

\usepackage[utf8]{inputenc} 
\usepackage[T1]{fontenc}    
\usepackage{amsfonts}       
\usepackage{booktabs}       
\usepackage{hyperref}       
\usepackage{url}            
\usepackage{microtype}      
\usepackage{nicefrac}       
\usepackage{paralist}       
\usepackage{siunitx}        
\usepackage{amsmath,amssymb}
\usepackage{mymacros}
\usepackage{graphicx}
\usepackage[numbers]{natbib}
\graphicspath{{./figs/}}

\title{
  Sheaf Neural Networks
}

\author{%
  Jakob Hansen\thanks{Equal contribution} \\
  Department of Mathematics \\
  The Ohio State University \\
  \texttt{hansen.612@osu.edu}
   \And
   Thomas Gebhart$^*$ \\
   Department of Computer Science \\
   University of Minnesota \\
   \texttt{gebhart@umn.edu} \\
}

\begin{document}

\maketitle

\begin{abstract}
  We present a generalization of graph convolutional networks by generalizing the diffusion operation underlying this class of graph neural networks. These \introduce{sheaf neural networks} are based on the \introduce{sheaf Laplacian}, a generalization of the graph Laplacian that encodes additional relational structure parameterized by the underlying graph. The sheaf Laplacian and associated matrices provide an extended version of the diffusion operation in graph convolutional networks, providing a proper generalization for domains where relations between nodes are non-constant, asymmetric, and varying in dimension. We show that the resulting sheaf neural networks can outperform graph convolutional networks in domains where relations between nodes are asymmetric and signed.
\end{abstract}

\section{Introduction}
Graph neural networks are a class of generalized neural network architectures 
that take as input relational data and learn to classify the input graph or 
its nodes. Because this relational data lacks a constant local 
Euclidean structure, the definition of a local convolution-type operator 
through which weight sharing may be achieved is a non-trivial architectural 
challenge. A natural approach, inspired by the theory of graph signal processing \cite{ortega_graph_2018} is to define 
convolution via the graph Laplacian or adjacency matrices 
\cite{bruna2013spectral, defferrard2016convolutional, kipf2017semi} so that the 
layer-wise convolution operation acts as neighborhood averaging followed by 
propagation of the updated signal, akin to message-passing diffusion with an 
additional element-wise non-linearity \cite{wu2019simplifying}. The association with convolution comes from taking the eigenvectors of the Laplacian or adjacency matrix as a Fourier basis for signals on the graph; convolution is then defined as multiplication of signals in this spectral domain. Actually computing the spectral coefficients is computationally intensive, so it is common to instead parameterize convolutional filters by polynomials in the adjacency or Laplacian operators. Indeed, using a polynomial of degree 1 works well in many situations \cite{kipf2017semi} and provides a computationally efficient architecture. These convolution operations, broadly construed, 
define a class of graph neural networks which we refer to as Graph Convolutional Networks (GCN).

We introduce a generalization of this GCN framework by leveraging a natural 
extension of the graph Laplacian called the sheaf Laplacian \cite{hansen_toward_2019}. 
Sheaf Laplacians form a class of local linear operators on a graph that respect 
the topological and algebraic structure of data associated to nodes. This generalization 
allows more complex relationships between nodes to be expressed while maintaining a consistent notion of local averaging and diffusion. After 
introducing cellular sheaves, sheaf Laplacians, and sheaf diffusion operators, we 
define sheaf neural networks and show that they are a useful generalization of GCNs 
in domains where the relational interaction between nodes across edges is non-constant. 
We validate this claim by comparing the performance of sheaf neural networks 
to GCNs on synthetic, semisupervised classification 
problems defined over signed graphs and find that sheaf neural networks significantly outperform the Kipf-Welling GCNs in this domain.

\section{Cellular sheaves}

A \introduce{cellular sheaf} is an algebraic-topological structure associated with a graph
that attaches spaces of data to nodes and edges. To be precise, a cellular
sheaf $\Fc$ on an undirected graph $G$ is given by specifying
\begin{itemize}
\item
  a vector space $\Fc(v)$ for each vertex $v$ of $G$
\item
  a vector space $\Fc(e)$ for each edge $e$ of $G$, and
\item
  a linear map $\Fc_{v \face e} : \Fc(v) \to \Fc(e)$ for each incident vertex-edge
  pair $v \face e$ of $G$.
\end{itemize}
The sheaf structure assigns spaces of data to vertices and edges,
and specifies consistency constraints for this data. For an edge $e$ between
vertices $u$ and $v$, we say that a choice of
data $x_v \in \Fc(v)$, $x_u \in \Fc(u)$ is \introduce{consistent} over $e$ if $\Fc_{v \face e} x_v = \Fc_{u \face e} x_u$. The space of data associated with all vertices of $G$ is
denoted $C^0(G;\Fc)$ and called the space of $0$-cochains valued in $\Fc$. One thinks of $C^0(G;\Fc)$ as a space of \introduce{signals} on the vertices of $G$, where the value of a signal at a vertex $v$ lives in the vector space $\Fc(v)$. Each
edge of $G$ imposes a constraint on $C^0(G;\Fc)$ by restricting the space
associated with its two incident vertices. The subspace of $C^0(G;\Fc)$
satisfying all these constraints is the space of \introduce{global sections} of
$\Fc$, and is denoted $H^0(G;\Fc)$.

There is likewise a space of signals associated with edges, denoted $C^1(G;\Fc)$.
The space of global sections $H^0(G;\Fc)$ is the kernel of a linear map $\delta:
C^0(G;\Fc) \to C^1(G;\Fc)$. This map is called the \introduce{coboundary}, and,
given an arbitrary choice of orientation on the edges of the graph, may be computed by 
\[(\delta x)_e = \Fc_{v \face e} x_v - \Fc_{u \face e} x_u\]
for each oriented edge $e = u \to v$. Clearly, if $x \in \ker \delta$, then
$\Fc_{v \face e} x_v = \Fc_{u \face e} x_u$ for every edge $e = u \sim v$. From
the coboundary operator we may construct the \introduce{sheaf Laplacian} $L_\Fc
= \delta^T\delta$, which is a positive semidefinite linear operator on
$C^0(G;\Fc)$ with kernel $H^0(G;\Fc)$, and does not depend on the orientation chosen for the edges of $G$.

A cellular sheaf operates as an extension of the structure of a graph. Rather
than simply recording connections between nodes, it specifies relationships
between data associated with those nodes. Standard graph-theoretic constructions
like Laplacians and diffusion operators implicitly work with the
\introduce{constant sheaf} on a graph: the sheaf $\constshf{\R}$ with all stalks
$\R$ and all restriction maps the identity. This is a very simple sort of
relationship between nodes, and can be greatly generalized in the sheaf setting. For instance, a sheaf can easily represent a signed graph by changing the sign of one restriction map of the constant sheaf for each negatively signed edge. Even more general relationships between nodes can be expressed, especially as stalks increase in dimension, resulting in such operators as connection Laplacians \cite{singer_vector_2012} and matrix-weighted Laplacians \cite{tuna_synchronization_2016}. 
In this paper we will simplify the constructions by only considering sheaves with constant-dimensional vertex stalks $\R^k$. More information on cellular sheaves and their Laplacians may be found in \cite{curry_sheaves_2014,hansen_toward_2019}, including extensions to higher-dimensional base spaces and additional topological context.

\subsection{Sheaf diffusion operators}

By an $r$-step \textit{local operator} associated to $\Fc$ we mean any linear
operator $D_\Fc$ on $C^0(G;\Fc)$ which is local with respect to $G$, in the sense
that $(D_\Fc x)_v$ depends only on $x_u$ for nodes $u$ in the $r$-step
neighborhood of $v$. We will suggestively call such an operator a \textit{sheaf diffusion operator} if it has nice
properties with respect to the algebraic structure of $\Fc$---for instance, if
sections of $\Fc$ form an eigenspace of $D_\Fc$. One such operator is the sheaf
Laplacian $L_\Fc = \delta^T\delta$. This is a $1$-step diffusion operator whose
zero eigenspace consists of sections of $\Fc$. For an appropriately chosen
$\alpha$, the operator $H_\Fc^\alpha = I - \alpha L_\Fc$ will have $2$-norm 1
and have $H^0(G;\Fc)$ as the eigenspace corresponding to the eigenvalue $1$.
There is also a normalized form of the sheaf Laplacian, $\tilde L_\Fc =
D^{-1/2}L_\Fc D^{-1/2}$, where $D$ is the block diagonal of $L_\Fc$. It amounts to the Laplacian for a version of $\Fc$ with reparameterized stalks, and its eigenvalues lie between $0$ and $2$.
This means that no scaling factor is necessary to construct a stable diffusion operator
$\tilde H_\Fc = I - \tilde{L}_\Fc$.
Diffusion operators depending on larger neighborhoods may be constructed from
powers of these operators. For any $r$, $(L_\Fc)^r$ and $(H_\Fc^\alpha)^r$ are
$r$-step sheaf diffusion operators.

From the standpoint of graph signal processing~\cite{ortega_graph_2018}, we can consider a sheaf diffusion operator $D_\Fc$ as a shift operator generating convolution-like filters for signals in $C^0(G;\Fc)$. We can define the convolution of two signals $x,y \in C^0(G;\Fc)$ by taking an eigendecomposition $D_\Fc = S\Lambda S^{-1}$ and letting $x * y = S(S^{-1} x \circ S^{-1} y)$, i.e. by pointwise multiplication in the eigenbasis of $D_\Fc$. By standard algebraic arguments, for a fixed $y \in C^0(G;\Fc)$ one can find a polynomial $P = a_0 I + a_1 D_\Fc + \cdots + a_N D_\Fc^N$ in $D_\Fc$ such that $x * y = Px$. Thus we can parameterize sheaf convolutions by polynomials in $D_\Fc$, without having to compute an eigendecomposition. This parameterization also gives natural bases for restricting the dimension of the space of convolutional filters. Indeed, in the context of graph convolutional networks, it is common to use a a degree-1 polynomial, since this significantly reduces the number of parameters to learn.

\section{Sheaf neural networks}
Local graph operators have been used to construct GCNs. These typically rely on applying one or more graph diffusion operators to the features on each layer and then taking a node-independent linear combination of features. A \introduce{sheaf neural network} is an extension of this framework.

Suppose that each node has $N_{\text{feat}}$ $k$-dimensional features. We write
the array of all features for all nodes as an $N_vk \times N_{\text{feat}}$ matrix $X$. A single
forward diffusion step applies a sheaf diffusion operator $D_\Fc$ for some sheaf $\Fc$ with $k$-dimensional stalks to the feature matrix $X$; that is, it calculates
$D_\Fc X$, producing a new set of $N_{\text{feat}}$ $k$-dimensional features for each
node. These features are neighborhood mixtures of the previous features,
computed in a way that is consistent with the algebraic structure of the sheaf.
To construct a neural network layer, we also apply a linear operator to each
nodewise feature vector. When $k=1$ this is realized by a right multiplication
$X A$ for some matrix $A$ with $N_{\text{feat}}$ rows.
For $k > 1$ one must also choose some $k\times k$ matrix $B$ and multiply it on the left to each block row of $X$. That is, the total operation on $X$ is $(I \otimes B) X A$, where $I \otimes B$ is the Kronecker product of $B$ with the $N_v \times N_v$ identity matrix; this amounts to a block diagonal matrix with copies of $B$ on the diagonal. After these linear transformations, we apply a stalkwise nonlinearity $\rho: \R^k \to \R^k$.

Combining these, a sheaf neural network layer for a sheaf $\Fc$ on $G$ has
hyperparameters $N_{\text{feat}}^{\text{in}}$, $N_{\text{feat}}^{\text{out}}$,
and $D_{\Fc}$, a diffusion operator associated with $\Fc$, together with a
stalkwise nonlinearity $\rho:\R^k \to \R^k$. Its learnable parameters are
the $N_{\text{feat}}^{\text{in}} \times N_{\text{feat}}^{\text{out}}$ matrix $A$
and the $k \times k$ matrix $B$. The action of this layer
$\text{SheafConv}(A,B)$ on a matrix $X$ of nodewise features is
\[\text{SheafConv}(A,B)(X) = \rho\left( D_\Fc (I \otimes B) X A \right).\]


We can use multiple diffusion operators in parallel by taking a 
concatenation or learnable linear combination of sheaf neural network layers (without the nonlinearity $\rho$). 
This might be useful for a task where we expect some behavior to be driven 
purely by connectivity, while other behavior depends on the relationships 
between nodes, and can also be used to combine powers of a diffusion operator to parameterize higher-order sheaf convolutional filters.

\section{Semisupervised classification}\label{sec:semisupervised_classification}
One problem for which graph-based neural networks are frequently used is the
semisupervised classification problem. In this problem, a network of objects is
given, together with $N_{\text{feat}}^{\text{in}}$ nodewise features and $N_{\text{class}}$ class labels associated with a (typically small) subset of the nodes. 
The goal is to impute class labels for the remaining nodes. For our sheaf-based approach to  meaningfully differ from a simpler graph neural network, we need an appropriate sheaf from which to construct a diffusion operator. 
Unfortunately, many popular graph classification datasets 
do not admit obvious sheaf 
Laplacian operators that are meaningfully different from the standard graph Laplacian.
Due to the lack of appropriate benchmark datasets, we illustrate the potential for sheaf-based neural networks using a synthetic family of semisupervised node classification problems over signed graphs. This allows us to isolate the contribution of a correctly-chosen sheaf diffusion operator in an appropriate setting.

To generate the synthetic node classification problem, we begin with a set of $N_v$ nodes, and assign each node an intrinsic feature vector $x_v \in \R^{N_\text{intrinsic}}$, sampled from a standard normal distribution.
We choose a classification vector $c \in \mathbb{R}^{N_{\text{intrinsic}}}$, and for each node $v$ choose a class $C_v \in \{\pm 1\}$ by $C_v = \text{sign}(\ip{c,x_v})$. 
We consider two sets of input features: one linear in the intrinsic features and one nonlinear. The linear features are a noisy random 
linear transformation of the intrinsic features: $x_v^{\text{in}} = P x_v + \epsilon_v$, 
where $P$ is a random $N^{\text{in}}_{\text{feat}} \times N_{\text{intrinsic}}$ matrix with independent standard Gaussian entries and
$\epsilon$ is a noise term obtained by independent samples from a normal distribution with mean zero and variance $\sigma_{\text{feat}}^2$. The nonlinear feature regime extends the map producing the linear features to a random 2-layer fully connected neural network: $x_v^{\text{in}} = P_2 \text{ReLU}(P_1 x_v + \epsilon_v)$, where $P_1$ and $P_2$ are random matrices with independent standard normal entries. 
We include this nonlinear regime to provide a stronger inferential challenge. 
It helps evaluate the potential performance of the SheafNN architecture for problems where the features and classes are not linearly related.

We generate a noisy graph from the intrinsic features by computing signed weights $w_{uv} = \ip{x_u,x_v} + \epsilon_{uv}$ and imputing a signed edge with weight $w_{uv}$ when $\abs{w_{uv}} > \tau$. Here $\epsilon_{uv}$ is drawn from a mean-zero Gaussian with variance $\sigma_w^2$. From this signed graph data, we construct a cellular sheaf $\Fc$ with all stalks $\R$, where the restriction maps $\Fc_{v \face e}$ and $\Fc_{u \face e}$ for an edge $e = u \sim v$ are multiplication by $\pm\sqrt{\abs{w_{uv}}}$. The signs are determined by the parity of the edge: for a positively signed edge $\Fc_{v \face e}$ and $\Fc_{u \face e}$ have the same sign, and for a negatively signed edge they have the opposite sign. The diffusion operator used is then $D_\Fc = H_\Fc^{1/d_{\max}} = I - \frac{1}{d_{\max}} L_\Fc$, where $d_{\max}$ is the maximum degree of the underlying graph.


For both the linear and non-linear noise regimes, we initialize two diffusion-based 
sheaf neural networks. The first architecture is three layers with 32 hidden dimensions in each layer (SheafNN-32). The second architecture is four layers with 16 hidden dimensions in each layer (SheafNN-16). For comparison, we also initialize two traditional diffusion-based graph convolutional networks according to \citep{kipf2017semi}
with equivalent hidden dimensions and number of layers (GCN-32, GCN-16). All models have ReLU activations. For each model type, we instantiate five random networks of 5000 nodes and simulate noisy intrinsic data according to the procedure in the previous paragraph. We set $N_{\text{intrinsic}} = 25, \tau=0.5, N_{\text{feat}}^{\text{in}} = 32$ and train each model according to this binary classification task, training on 75\% of nodes and determining test performance on the rest. We train each model for $1000$ epochs using Adam gradient optimization \cite{kingma2014adam} with learning rate $0.001$. 

\begin{figure}[htbp]
    \centering
    \includegraphics[width=\textwidth]{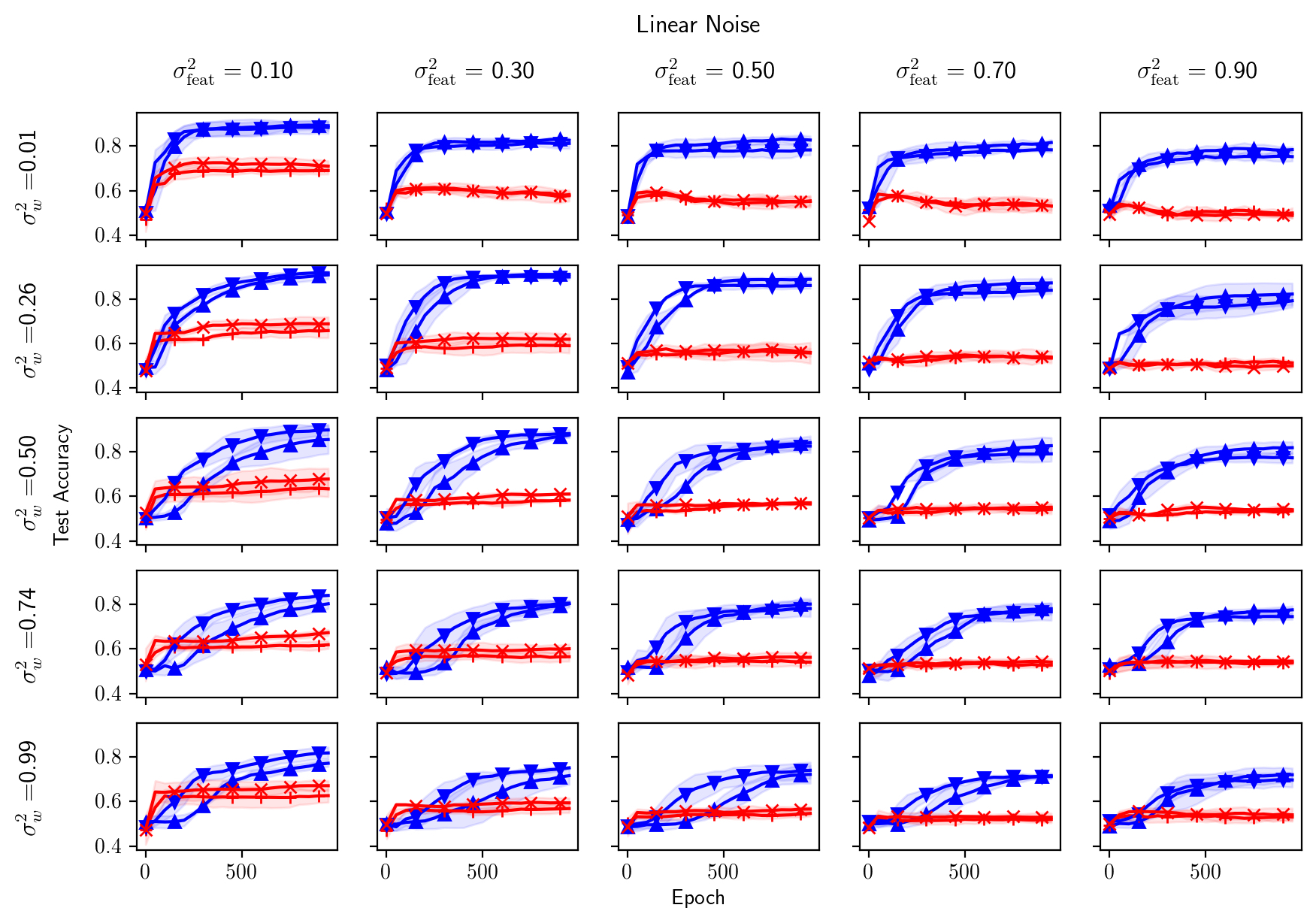}
    \includegraphics[width=\textwidth]{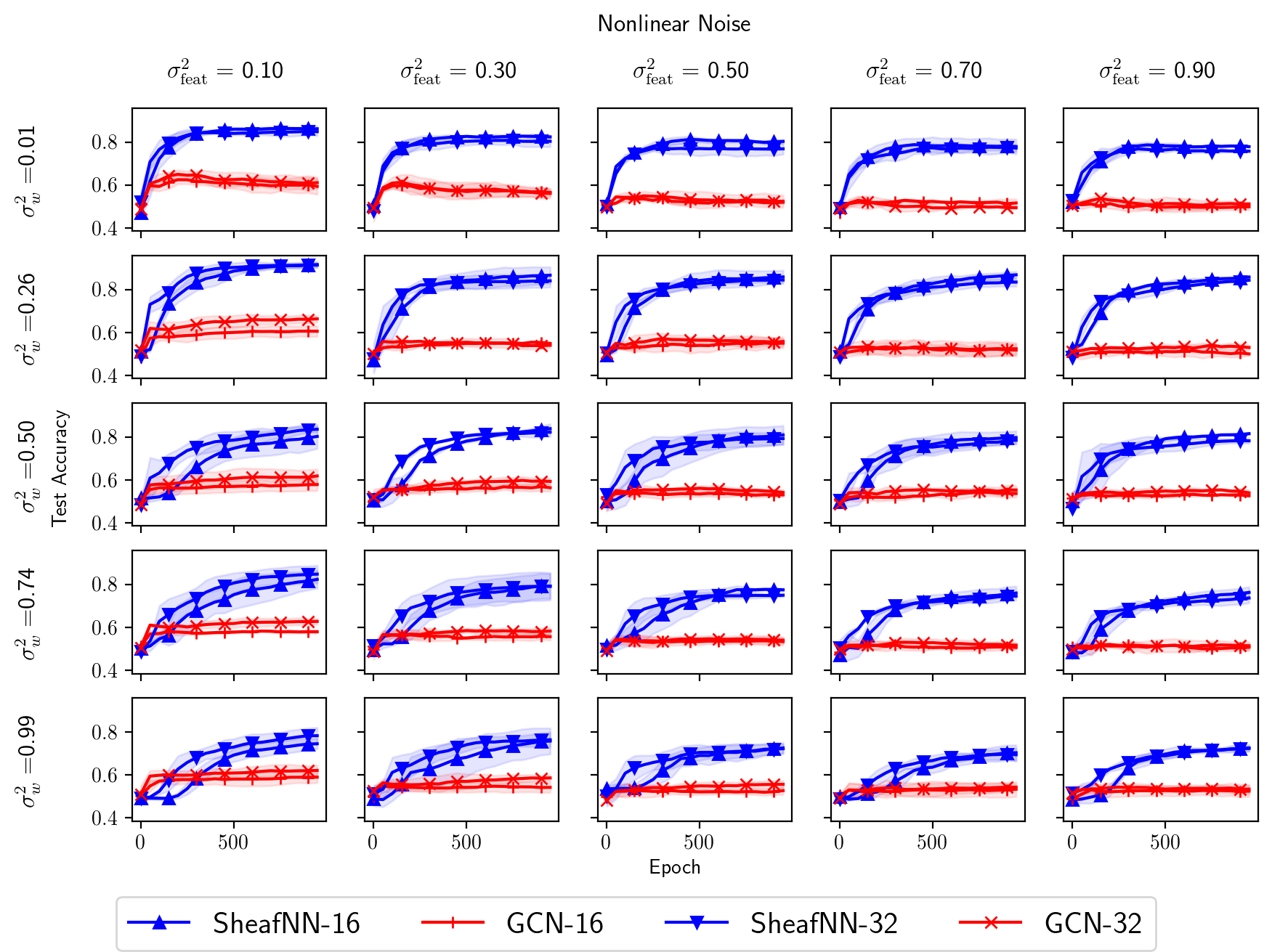}
    \caption{\textbf{Sheaf neural networks outperform graph convolutional networks on signed graphs.} Comparison of SheafNN and GCN models on the semisupervised classification task described in Section 
    \ref{sec:semisupervised_classification}. Lines plotted are the mean across five random graph trials, with the standard deviation in error bars. Rows correspond to noise levels for edge weights, while columns correspond to noise levels for input features.}
    \label{fig:sheafnn-gcn}
\end{figure}

Figure \ref{fig:sheafnn-gcn} plots the performance of these two models in both the linear and nonlinear feature generation regimes. Clearly, the SheafNN variants nearly all outperform the GCN variants under a wide range of feature noise levels ($\sigma_{\text{feat}}^2$) and weight noise levels ($\sigma_w^2$). The GCN models appear to saturate in accuracy slightly above chance. This is expected as their underlying diffusion operation does not respect the signed nature of the graph. Increasing the weight noise level has the effect of decreasing performance and slowing training convergence. This is expected as increasing the noise across edge weights blurs the intrinsic similarity between nodes which has the effect of promoting false relations between nodes, masking true relations between intrinsically similar nodes, or flipping the intrinsic relationship altogether. Increasing feature noise uniformly dampens the maximal classification accuracy across all models as the underlying signal becomes distorted.

\section{Discussion}
The sheaf Laplacian is as a proper generalization of the graph Laplacian in defining diffusion operations for domains where relations between nodes are non-constant, asymmetric, and varying in dimension. This sheaf diffusion operator acts as a drop-in replacement for the graph diffusion operator and outperforms the standard graph convolutional model in semisupervised node classification over signed graphs. There are many avenues for future research related to sheaf neural networks. 

As mentioned in Section \ref{sec:semisupervised_classification}, most standard graph datasets do not offer obvious sheaf structures to leverage, which makes applying sheaf neural networks a more difficult task. We view this fact as a reflection of the early stage of development of graph classification and processing. There are many relational processes in nature that are most naturally modeled using asymmetric, heterogeneous relations. As datasets related to these processes emerge, we expect sheaf neural networks to outperform related algorithms defined using adjacency-based diffusion operators. 
Another possible avenue of application is to learn the sheaf structure from the dataset, as in \cite{hansen_learning_2019}. 
One exciting possibility is the combination of these approaches, where the sheaf structure is learned simultaneously with the solution to a classification or regression task.

Finally, we note that we have merely scratched the surface of cellular sheaf theory in the above sections. Numerous other ideas from cellular sheaf theory---sheaf morphisms, approximations, and pushforward/pullback operations---could offer even greater flexibility in analyzing graph datasets.

\bibliographystyle{plain}
\bibliography{sheafspectra}

\end{document}